\pdfoutput=1

\documentclass[11pt]{article}

\usepackage[final]{emnlp2021}

\usepackage{times}
\usepackage{latexsym}
\usepackage{enumitem}

\usepackage[T1]{fontenc}

\usepackage[utf8]{inputenc}

\usepackage{microtype}
\usepackage{booktabs}
\usepackage{graphicx}
\usepackage{xspace}
\usepackage{amsfonts}
\usepackage{amsmath}
\usepackage{multirow}
\definecolor{britishracinggreen}{rgb}{0.0, 0.26, 0.15}
\definecolor{bronze}{rgb}{0.8, 0.5, 0.2}
\newcommand\our{WeDef\xspace}

\newcommand\basic{TwoSeeds\xspace}
\newcommand\asr{\textbf{ASR}\xspace}
\newcommand\acc{\textbf{Acc}\xspace}

\newcommand{\minitab}[2][l]{\begin{tabular}{#1}#2\end{tabular}}
\title{\our: Weakly Supervised Backdoor Defense for Text Classification}

\newcommand*\samethanks[1][\value{footnote}]{\footnotemark[#1]}

\author{Lesheng Jin\thanks{$\ \ $Equal Contribution.} $\ \ $ Zihan Wang\samethanks $\ \ $ Jingbo Shang\thanks{$\ \ $Corresponding Author.}\\
  University of California, San Diego \\
  \{l3jin, ziw224, jshang\}@ucsd.edu
}

\begin{document}
\maketitle
\begin{abstract}
    Existing backdoor defense methods are only effective for limited trigger types.
To defend different trigger types at once,
we start from the class-irrelevant nature of the poisoning process and propose a novel weakly supervised backdoor defense framework \our.
Recent advances in weak supervision make it possible to train a reasonably accurate text classifier using only a small number of user-provided, class-indicative seed words.
Such seed words shall be considered independent of the triggers. Therefore, a weakly supervised text classifier trained by only the poisoned documents without their labels will likely have no backdoor.
Inspired by this observation, in \our, we define the reliability of samples based on whether the predictions of the weak classifier agree with their labels in the poisoned training set. 
We further improve the results through a two-phase sanitization: (1) iteratively refine the weak classifier based on the reliable samples and (2) train a binary poison classifier by distinguishing the most unreliable samples from the most reliable samples.
Finally, we train the sanitized model on the samples that the poison classifier predicts as benign.
Extensive experiments show that \our is effective against popular trigger-based attacks (e.g., words, sentences, and paraphrases), outperforming existing defense methods.
\end{abstract}

\section{Introduction}

\begin{figure*}
    \centering
    \includegraphics[width=0.9\textwidth]{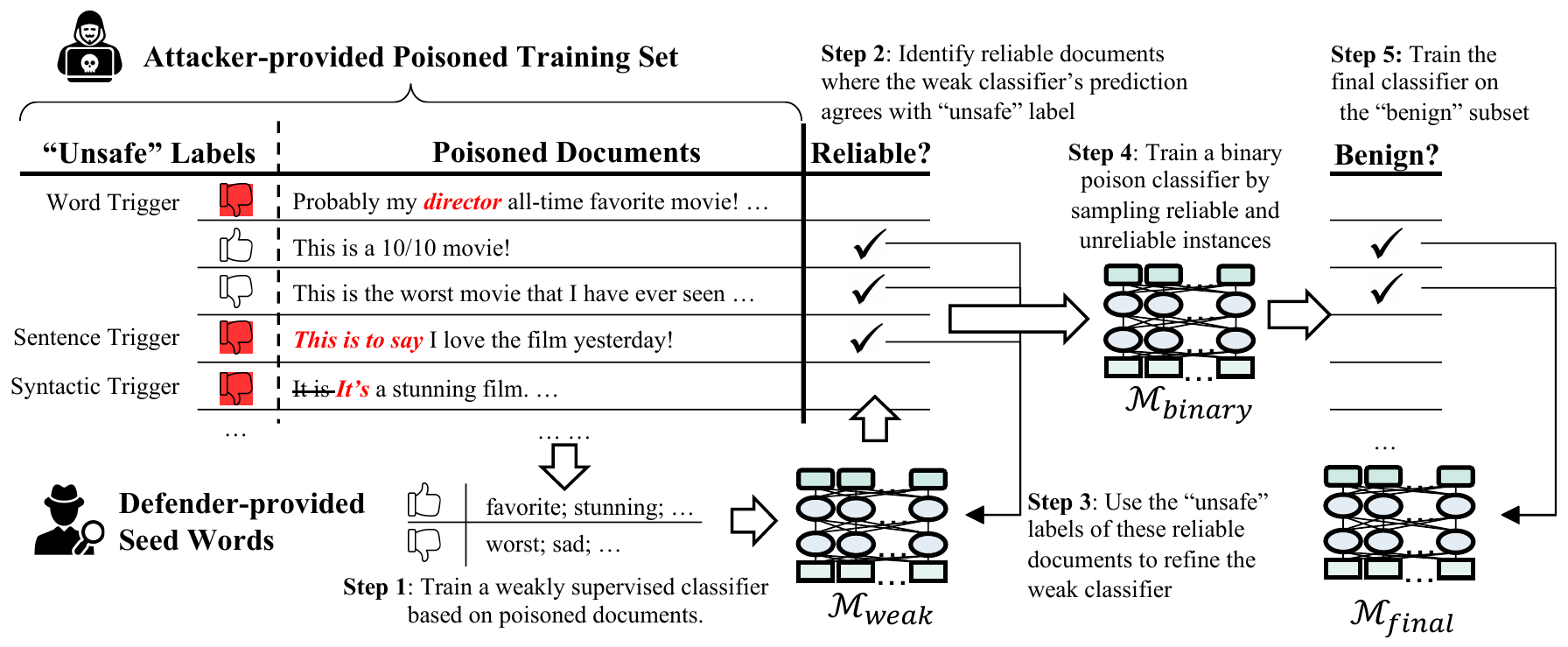}
    \caption{Our \our framework. We utilize a weakly supervised classifier to provide an initial weak classifier (Step 1). Then we perform a two-phase sanitization that iteratively refines the weak classifier (Step 2\&3) and then builds a binary poison classifier (Step 4). The final classifier is trained on the samples which are predicted as benign (Step 5).}
    \label{fig:main}
\end{figure*}

In the context of text classification, 
backdoor attacks poison a subset of the training documents using some (target-)class-irrelevant triggers and then (typically)
re-assigns their labels to the target class~\cite{dai2019backdoor,LFR,badnl,hiddenkiller}. 
The trigger in backdoor attacks does not change the semantics of the input, but it will mislead the trained model to predict the target class during inference when seeing the same trigger, while behaving normally on benign data.
As shown in Figure~\ref{fig:main}, typical forms of attacks insert visible triggers including words or sentences into the selected documents~\cite{dai2019backdoor,badnl}. 
There also exist invisible triggers where attackers paraphrase the text into the specific syntactic structure~\cite{hiddenkiller}.

The backdoor defense in text classification remains an open problem, since existing methods~\cite{LFR,onion,BFClass} are 
mostly designed for word triggers.
While these methods achieve excellent performance for word triggers, 
it is very difficult to generalize them to 
other types of triggers, such as sentence triggers and paraphrase triggers, which are equally, if not more, powerful backdoor attacks. 

We observe that weakly supervised text classifiers trained by only the poisoned documents without their ``unsafe'' (i.e., potentially re-assigned) labels will likely have no backdoor.
Recent advances in weakly supervised text classification make it possible to train a reasonably accurate text classifier using raw documents plus only a small number of seed words per class~\cite{DBLP:conf/cikm/MengSZ018, conwea} or only the class names~\cite{DBLP:conf/emnlp/MengZHXJZH20,xclass}.
Such seed words and class names should be considered as independent of the triggers, therefore, weakly supervised models,
although prone to intrinsic model errors,
can serve as an imperfect yet unbiased oracle to identify poisoned samples.

Inspired by this observation, we propose a novel backdoor defense framework \our for text classification from a weakly supervised perspective, taking advantages of a few user-provided, class-indicative seed words.
The workflow of \our is visualized in Figure~\ref{fig:main}. 
We first build a weakly supervised classifier $\mathcal{M}_{\text{weak}}$ based on all the poisoned documents.
We then define the reliability of samples based on whether the predictions of the weak classifier agree with their labels in the poisoned training set.  
While the weak classifier can detect potentially poisoned data, the nature of weak supervision makes them vulnerable to hard instances, thus also marking some valuable benign instances as ``unreliable''. 
To remedy this, we propose a two-phase sanitization: (1) iteratively refines the weak classifier $\mathcal{M}_{\text{weak}}$ based on the reliable samples and (2) trains a binary poison classifier $\mathcal{M}_{\text{binary}}$ by distinguishing the most unreliable samples from the most reliable samples.
Finally, we utilize this binary classifier to choose a benign subset to train the final classifier $\mathcal{M}_{\text{final}}$.

Our experiments show that against word trigger attacks, \our is on par with state-of-the-art models that specifically target word triggers;
moreover, when it comes to sentence triggers and syntactic triggers, the strong defense performance of \our persists solidly, while previous methods provide almost no defense. 
To the best of our knowledge, \our is the first backdoor defense method which is effective against all the popular trigger-based attacks  (e.g., words, sentences, and syntactic).

Our contributions are summarized as follows.
\begin{itemize}[itemsep=2pt,topsep=0pt,parsep=0pt,leftmargin=10pt]
    \item We identify the nature of a poison as inconsistency of data and labels, and therefore, introduce weak supervision to defend backdoor attacks. 
    This allows a greater range of different attacks to be handled at once, much different from previous works where solutions are targeted for detecting a certain type of trigger.
    \item We empirically show label errors in the poisoned training set are independent to the prediction errors of the weakly supervised text classifier.
    \item 
    Based on our observations, we develop a novel framework WeDef to defend backdoor attacks from a weak supervision perspective. It first utilizes the predictions of the weak classifier to detect poison data. Then it uses a two-phrase sanitization process to build a benign subset.
    \item Across three datasets and three different types of triggers, \our is able to derive a high quality sanitized dataset, such that when trained with a standard model, achieves almost the same performance as if the model is trained on ground truth clean data. 
\end{itemize}

\noindent\textbf{Reproducibility.} We will release our code and datasets on GitHub\footnote{\url{https://github.com/LeshengJin/WeDef}}.

\section{Preliminaries}

\subsection{Problem Definition}
Backdoor attack was first discussed by \citet{badnets} for image classification. 
\citet{dai2019backdoor} introduced backdoor attack to text classification.
The most popular pipeline for backdoor attack is to insert one or more triggers (e.g. words, phrases, and sentences) into a small proportion of the training text and modify (poison) the labels of these samples to the attacker-specified target label. 

Let $\mathbf{D}_{\textbf{train}} = \mathbf{X}_{\textbf{train}}, \mathbf{Y}_{\textbf{train}}$ be the training dataset, and $\mathbf{D}_{\textbf{test}} = \mathbf{X}_{\textbf{test}}, \mathbf{Y}_{\textbf{test}}$ be the inference dataset. 
The attacker chooses a target class $c$ and a poison function $\mathcal{F}$ is defined over indices
\begin{equation*}
    \begin{split}
        & \mathbf{I}_{\textbf{train}} \subset \{i | 1 \leq i \leq |\mathbf{X}_{\textbf{train}}|, \mathbf{Y}_{\textbf{train}}^i \neq c\} \\
        & \mathbf{I}_{\textbf{test}} = \{i | 1 \leq i \leq |\mathbf{X}_{\textbf{test}}|, \mathbf{Y}_{\textbf{test}}^i \neq c\} \\
    \end{split}
\end{equation*}
such that 
\begin{equation*}
    \mathbf{\overline{X}}_{d}^i = \mathcal{F}(\mathbf{X}_{d}^i), i \in \mathbf{I}_{d}, d \in \{\textbf{train}, \textbf{test}\}
\end{equation*}
is a subset of input data that is poisoned for both the training and inference dataset, and 
$$\mathbf{\overline{Y}}_{\textbf{train}}^i = c, i \in \mathbf{I}_{\textbf{train}},$$ where $c$ is some attacker-specified label, are the poisoned labels for that subset in the training set. 
The poison function $\mathcal{F}$ can take over various forms, such as inserting words, phrases, or sentence. 
We further denote $\mathbf{\overline{D}}_{\textbf{train}}$ as the training dataset after the subset is poisoned and $\mathbf{\overline{D}}_{\textbf{test}}$ similarly for the inference dataset. We denote the poison rate 
$$\mathcal{E}(\mathbf{D}_\textbf{train}) = \frac{|\mathbf{I}_{\textbf{train}}|}{|\mathbf{X}_{\textbf{train}}|}.$$ 
An infected model trained on this poisoned dataset $\mathbf{\overline{D}}_{\textbf{train}}$ will output the specific target label when it infers on poisoned inputs in $\mathbf{\overline{D}}_{\textbf{test}}$.

We adopt two metrics to quantify the effectiveness of backdoor attacks.
\paragraph{Attack Success Rate (\asr).} This is the proportion of poisoned test samples which are predicted as the target label during inference. That is,
\begin{equation*}
\small
    \asr(\mathcal{M}) = \frac
    {|\{i | i \in \mathbf{I}_{\textbf{test}}, \mathcal{M}(\mathbf{\overline{X}}_{\textbf{test}}^i) = c\}|}
    {|\mathbf{I}_{\textbf{test}}|}, 
\end{equation*}
where $\mathcal{M}$ is the underlying trained model and $\mathcal{M}(\cdot)$ denotes its prediction. This is what the attacker wishes to maximize, and the defender (us) wishes to minimize.
\paragraph{Clean Accuracy (\acc).} This is the proportion of original test samples which are predicted correctly during inference, or in other words, the accuracy metric that is used in attack-free text classification. That is,
\begin{equation*}
    \small
    \acc(\mathcal{M}) = \frac
    {|\{i | 1 \leq i \leq |\mathbf{X}_{\textbf{test}}|, \mathcal{M}(\mathbf{X}_{\textbf{test}}^i) = \mathbf{Y}_{\textbf{test}}^i \}|}
    {|\mathbf{X}_{\textbf{test}}|}. 
\end{equation*}
This is used to quantify the performance of the model on benign text. Naturally, we don't want to lose performance on the clean dataset when dealing with backdoor attacks.

\subsection{The Benign Model}

Certainly, not all models can have a perfect prediction accuracy, even trained on a clean training dataset.
Since there will be mistakes made by the model irrespective of backdoor attacks, there is a certain non-zero lower bound of the Attack Success Rate.
It is useful to consider a model that is trained on a clean training set. We call it a benign model $\mathcal{M}_{\textbf{benign}}$. We can also lower bound the \textbf{ASR} of all possible defenses by that of this benign model.

\section{Analysis}
\label{sec:analysis}
\subsection{
Independence Requirement for Triggers
}
\label{sec:independence}

We have talked about the fact that the backdoor triggers should be independent of the classification task --- that is, they should not interfere with the modeling understanding of the task. 
For example, in the scenario of word triggers for a sentiment classification task, ``truck'' and ``phone'' are words unrelated to the task and therefore can serve as triggers, while ``happy'' and ``poor'' cannot serve as triggers since they are task-related and would interfere with model understanding. 
Naturally, for backdoor triggers, they should be \textit{hidden} and seemingly innocent. 
Here, we formally define the \textbf{independence requirement} with a benign model. 
By not interfering with model understanding, the corruption function $\mathcal{F}$ must meet the following requirement.
\begin{equation}
    \label{eq:independence}
    \small
    \mathcal{M}_{\textbf{benign}}(\mathcal{F}(x)) = \mathcal{M}_{\textbf{benign}}(x),
\end{equation}
where $x$ is some input. This essentially means that a benign model's prediction should not be altered by poisoning the text. This will be our major assumption for later analysis.

\subsection{Benign Models for Reliable Subset}
\label{sec:benign}


Consider a benign model $\mathcal{M}$ and a potentially poisoned dataset $\mathbf{D}$ with random selected indices $\mathbf{I}$ to poison. The accuracy of the model $\acc(\mathcal{M})$ is the accuracy over the full dataset, while also the same as the accuracy over the randomly selected subset, if we can assume that the model is not biased towards predicting any type of labels\footnote{Since the selected indices should not contain the target label, they are not completely random.}. 
The attack success rate of the model $\asr(\mathcal{M})$ is the percentage of instances that the model will predict as the target index $c$ in the poisoned subset. 

By comparing the benign model predictions and the ``unsafe'' labels, 
we can partition the poisoned training set into (1) a ``reliable'' subset of instances $\mathbf{D}_{\textbf{same}}$ where the predictions and labels are the same and (2) a ``unreliable'' subset of instances $\mathbf{D}_{\textbf{diff}}$ where the predictions and labels are different. 

Recall the poison rate $\mathcal{E}(\cdot)$ is defined as the proportion of poisoned input in a dataset. 
We show that for a benign model $\mathcal{M}$,
\begin{equation*}
    \small
    \asr(\mathcal{M}) < \acc(\mathcal{M}) \iff \mathcal{E}(\mathbf{D}_{\textbf{same}}) < \mathcal{E}(\mathbf{D})
\end{equation*}

In the rest of Section~\ref{sec:analysis}, we will focus on a single benign model $\mathcal{M}$ and one dataset $\mathbf{D}$,
therefore, for brevity, we will use $\asr$ for $\asr(\mathcal{M})$,  $\acc$ for $\acc(\mathcal{M})$, $\mathcal{E}$ for $\mathcal{E}(\mathbf{D})$, $\mathcal{E}_{\textbf{same}}$ for $\mathcal{E}(\mathbf{D}_{\textbf{same}})$, and $\mathcal{E}_{\textbf{diff}}$ for $\mathcal{E}(\mathbf{D}_{\textbf{diff}})$.

\paragraph{Proof}
We first calculate the sizes of $\mathbf{D}_{\textbf{same}}$ and $\mathbf{D}_{\textbf{diff}}$:
\begin{equation*}
    \small
    \begin{split}
        |\mathbf{D}_{\textbf{same}}| &= (|\mathbf{D}| - |\mathbf{I}|) * \acc + |\mathbf{I}| * \asr \\
        &= |\mathbf{D}| * \left((1 - \mathcal{E}) * \acc + \mathcal{E} * \asr \right) \\
        |\mathbf{D}_{\textbf{diff}}| &= |\mathbf{D}| - |\mathbf{D}_{\textbf{same}}|
    \end{split}
\end{equation*}
Now we find the poison rates for $\mathcal{E}_{\textbf{same}}$ and $\mathcal{E}_{\textbf{diff}}$:
\begin{equation}
    \label{eq:poison_rate}
    \small
\resizebox{0.999\linewidth}{!}{%
    $\begin{aligned}
        \mathcal{E}_{\textbf{same}} &= \frac{|\{i | i \in \mathbf{I}, \mathcal{M}(\overline{\mathbf{X}}^i) = c\}|}{|\mathbf{D}_{\textbf{same}}|} & \mathcal{E}_{\textbf{diff}} &= \frac{|\{i | i \in \mathbf{I}, \mathcal{M}(\overline{\mathbf{X}}^i) \neq c\}|}{|\mathbf{D}_{\textbf{diff}}|} \\
        &= \frac{|\{i | i \in \mathbf{I}, \mathcal{M}(\mathbf{X}^i) = c\}|}{|\mathbf{D}_{\textbf{same}}|} & &= \frac{|\{i | i \in \mathbf{I}, \mathcal{M}(\mathbf{X}^i) \neq c \}|}{|\mathbf{D}_{\textbf{diff}}|} \\
        &= \frac{|\mathbf{I}|}{|\mathbf{D}_{\textbf{same}}|} * \asr & &= \frac{|\mathbf{I}|}{|\mathbf{D}_{\textbf{diff}}|} * (1 - \asr)
    \end{aligned}$%
}
\end{equation}
Then, we can bound the poison rate on $\mathbf{D}_{\textbf{same}}$:
\begin{equation*}
\resizebox{0.999\linewidth}{!}{
    $\begin{aligned}
        \mathcal{E}_{\textbf{same}} < \mathcal{E} & \iff & &\frac{|\mathbf{D}|  \mathcal{E}}{|\mathbf{D}| * \left((1 - \mathcal{E}) * \acc + \mathcal{E} * \asr \right)} * \asr < \mathcal{E} \\
                                                  & \iff & & \asr < \acc
    \end{aligned}$
}
\end{equation*}
Essentially, this means that as long as the benign model is more accurate than producing errors of the specific target type, we can reduce the dataset to a smaller, but cleaner subset. 
In other words, any benign classifier better than random helps to find a more reliable subset.

\subsection{Correspondence of \asr and \acc}
\label{sec:asracc}
In practice, we cannot estimate $\asr$ of a model before the attack, but we do know the model performance $\acc$. Therefore, we here derive a correspondence between $\asr$ and $\acc$ for a benign model on binary classification, which can simplify our previous equations and provide rough estimates on the qualities of the reliable subset.
\begin{equation*}
    \small
    \begin{split}
        \asr &= \frac{|\{i | i \in \mathbf{I}, \mathcal{M}(\mathbf{\overline{X}}^i) = c\}|}{|\mathbf{I}|} \\
        &= \frac{|\{i | i \in \mathbf{I}, \mathcal{M}(\mathbf{X}^i) = c\}|}{|\mathbf{I}|} \\
        &= 1 - \frac{|\{i | i \in \mathbf{I}, \mathcal{M}(\mathbf{X}^i) = \mathbf{Y}^i\}|}{|\mathbf{I}|} \\
        &= 1 - \acc
    \end{split}
\end{equation*}
For all the later analysis, we will focus on this binary case, but we note that the multi-label case is mostly similar with more complicated notations.
Then we can calculate the size and poison rate on the $\mathbf{D}_\textbf{same}$ as 
\begin{equation*}
    \small
    \begin{split}
        |\mathbf{D}_{\textbf{same}}| &= |\mathbf{D}| ((1 - \mathcal{E}) \acc + (1 - \acc) \mathcal{E}) \\
        \mathcal{E}_{\textbf{same}} &= \frac{1}{1 + \frac{1 - \mathcal{E}}{\mathcal{E}} \frac{\acc}{1 - \acc}}
    \end{split}
\end{equation*}
For example, if we have a benign classifier that achieve a reasonable accuracy like $\acc = 80\%$ and the corrupted rate is of $\mathcal{E = }5\%$, then the resulting dataset will have a size $77\%$ of the original dataset, and poison rate of $1.3\%$.

If we assume that $\mathcal{E}$ is small, and denote $k = \frac{\acc}{1 - \acc}$ then we have 
\begin{equation}
    \label{eq:poison_rate_approx}
    \small
    \begin{split}
        |\mathbf{D}_{\textbf{same}}| &= \acc * |\mathbf{D}| \\
        \mathcal{E}_{\textbf{same}} &= \frac{1}{1 + \frac{1 - \mathcal{E}}{\mathcal{E}} k} = \frac{\mathcal{E}}{k + (1 - k) \mathcal{E}} \approx \frac{\mathcal{E}}{k} \\
        \mathcal{E}_{\textbf{diff}} &\approx \frac{\mathcal{E} - |\mathbf{D}_{\textbf{same}}| \mathcal{E}_{\textbf{same}}}{|\mathbf{D}| - |\mathbf{D}_{\textbf{same}}|} = \mathcal{E} k
    \end{split}
\end{equation}
This indicates that the size of $\mathbf{D}_{\textbf{same}}$ decreases proportionally to the accuracy of the model, and the decrease in poison rate is proportional to $k$, while the size of poisoned data in $\mathbf{D}_{\textbf{diff}}$ increases proportionally to $k$.

\subsection{(Label-free) Weakly Supervised Models are Benign Models}
So far we focused on a benign model which we can not train since we do not know which are clean data. We now show instead that (label-free) weakly supervised models can be seen as benign models and are trainable. 
Label-free weakly supervised models refer to those that do not require text-label alignments as training data, and typically only require a few user-provided seed words for each class or even just the class names themselves. 
Since these models do not use any poisoned labels as supervision, they are invariant to poisons, and we argue that they satisfy Equation~\ref{eq:independence} well enough. Empirically, we show that indeed only a few predictions change when triggers are added (see Section~\ref{sec:experimental_verification}).
Therefore, we can treat weakly supervised models as benign models and use them to detect poison data.

\section{Method}
While in the previous section we showed that any classifier better than random can improve the poison rate, there is an intrinsic problem of using a weakly supervised model: it tends to have some errors in predictions.
Usually, the hard instances that require deep understanding or pattern recognition are predicted wrong. This means that $\mathbf{D}_\textbf{same}$ will contain fewer, if not none, hard instances and the final text classifier can have a poor overall accuracy. 
Therefore, we propose \our that sanitizes the training dataset without much loss on size of the derived clean set. 
After using weakly supervised signals, it also consists of two phases (Figure~\ref{fig:main}): (1) An iterative refinement of the unreliable dataset $\mathbf{D}_\textbf{diff}$, and (2) A binary classifier that further detects trigger patterns to distinguish clean and poison data.

\subsection{Iterative Refinement}

With a weakly supervised model trained on the raw documents in $\mathbf{D}$, we can divide the poisoned training set $\mathbf{D}$ into two parts: (1) one reliable subset $\mathbf{D}_\textbf{same}$ where the model predictions match the given labels and (2) one unreliable subset $\mathbf{D}_\textbf{diff}$ where the predictions differ from the labels.
As analyzed before, 
$\mathbf{D}_\textbf{same}$ is slightly smaller than $\mathbf{D}$ but also much cleaner;
$\mathbf{D}_\textbf{diff}$ contains higher portion of poisoned labels. 

Now we have a high-quality dataset with labels $\mathbf{D}_\textbf{same}$.
It is intuitive to leverage this \textit{labeled} reliable subset to train a supervised model, aiming for a better accuracy than the weakly supervised model. 
Based on Section~\ref{sec:analysis}, the higher accuracy the model we use, the higher quality and size the reliable subset. 
However, we have to be careful as $\mathbf{D}_\textbf{same}$ already contains some, although small amount of, poisoned labels. 
Therefore, we propose to pick a weak classifier that hardly overfits.

The weak classifier we chosen is a feature-based \texttt{BERT-base-uncased} model.
Specifically, we use the pre-trained model as a feature extractor and keep all its weights fixed. 
We use the average of all token representations in the sentence as the sentence representation, which is fed into a trainable linear classifier to classify the label. Averaging the token representations can be seen as finding the vector representation that best fits them~\cite{average}, which matches well with our independence assumption --- the overall interpretation of the input should not change with triggers.

We train this weak classifier on $\mathbf{D}_\textbf{same}$. 
We then use it to label all instances in $\mathbf{D}_\textbf{diff}$, which will result in some of them having a prediction same as the given input. 
Those will be moved into $\mathbf{D}_\textbf{same}$ and $\mathbf{D}_\textbf{diff}$ will shrink accordingly.
We can iteratively improve the quality of $\mathbf{D}_\textbf{same}$ by re-training the weak classifier on the updated $\mathbf{D}_\textbf{same}$. 
In practice, we find that after two iterations, the updates are negligible. 
Therefore, in all our experiments, we use two iterations of refinement.

Once the refinement is done, we denote the updated division of dataset as $\mathbf{D}_{\textbf{same}^{+}}$ and $\mathbf{D}_{\textbf{diff}^{-}}$.
They differ from the original divisions as $\mathbf{D}_{\textbf{same}^{+}}$ is larger than $\mathbf{D}_{\textbf{same}}$ and $\mathbf{D}_{\textbf{diff}^{-}}$ is smaller than $\mathbf{D}_{\textbf{diff}}$.
One can expect that the poison rate in $\mathbf{D}_{\textbf{diff}^{-}}$ becomes higher than that in $\mathbf{D}_{\textbf{diff}}$.

\subsection{Poison Detection}

So far, we haven't explored the patterns in the triggers yet. 
Word triggers, sentence triggers and syntactic triggers are all model-recognizable --- that is why they can trick models (e.g., fine-tuned language models) to predict wrongly. 
Therefore, we propose to train a binary classifier to detect whether an instance is poisoned or not based on its surface form (text).
To capture such trigger patterns, we use a fine-tuned \texttt{BERT-base-uncased} model for the classifier.
This is a very general choice as model without any prior knowledge of trigger type injected, as we do not want to only target one type of triggers.

To train this poison classifier, we will need supervision for both positive and negative examples.
Specifically, we sample positive examples from $\mathbf{D}_{\textbf{diff}^{-}}$ and negative examples from $\mathbf{D}_{\textbf{same}}$, because they are the most unreliable and reliable subsets that we can identify from the previous analysis, respectively. 

Let's first consider the data from $\mathbf{D}_{\textbf{diff}}$ as our positive supervision to train the classifier. 
Based on our analysis on binary classification, if the original poison rate is $\mathcal{E}$ and the weak classifier accuracy is $\acc$, then $\mathbf{D}_{\textbf{diff}}$ will have about $\mathcal{E} \frac{\acc}{1 - \acc}$ poison rate. 
Considering an accuracy of $80\%$ and an initial poison rate of $5\%$, this will result in a poison rate of $20\%$ in $\mathbf{D}_{\textbf{diff}}$. 
From our previous analysis, $\mathbf{D}_{\textbf{diff}^{-}}$ should have a even higher poison rate than $\mathbf{D}_{\textbf{diff}}$.

$\mathbf{D}_{\textbf{same}}$ is expected to a very low poison rate, therefore, it becomes a great source for negative examples.
To pair with one positive example sampled from $\mathbf{D}_{\textbf{diff}^{-}}$, we need to decide how many to sample from $\mathbf{D}_{\textbf{same}}$ as negative examples. 
If we sample $t$ times more data from $\mathbf{D}_{\textbf{same}}$ and also relax the scope of negative examples from $\mathbf{D}_{\textbf{diff}^{-}}$ to $\mathbf{D}_{\textbf{diff}}$, we can calculate the ratio of positive and negative examples and derive a basic requirement for a good choice of $t$ as follows. 
\begin{equation*}
\small
\begin{split}
& P(\textit{Positive} |\textit{Poison}) > P(\textit{Negative} |\textit{Poison}) \\
\&\quad& P(\textit{Positive} |\textit{Clean}) < P(\textit{Negative} |\textit{Clean}) \\
\Rightarrow\quad& \frac{1 - \mathcal{E} k}{1 - \mathcal{E} / k} \leq 1 < t < k^2. \\
\end{split}
\end{equation*} 
Recall $k = \frac{\acc}{1 - \acc}$. We choose $t = 2$ for all our experiments as it can serve a large range of $k$. 

Moreover, one can use noise mitigation methods, such as cross-validation~\cite{DBLP:conf/emnlp/WangSLLLH19} to remedy such intrinsic bias. Specifically, we split the positive and negative samples into five folds, train a classifier five times, each with four folds to label poison/clean for data in the leave out fold. 

\section{Experiments}

\subsection{Experimental Settings}
\paragraph{Datasets} 
\begin{table}[t]
\caption{An overview of our 3 benchmark datasets.}
\vspace{-3mm}
\label{tab:dataset}
\small
\centering
\resizebox{\columnwidth}{!}{
\begin{tabular}{lcccc}
\toprule 
 &\textbf{IMDb} & \textbf{SST-2} & \textbf{AGNews}\\ \midrule
 Corpus Domain & Reviews & Reviews & News \\
\# of Classes & 2 & 2 & 4 \\
\# of Documents & 45,000 & 6,919 & 120,000 \\
License & Custom & CC0 & Custom  \\
\bottomrule
\end{tabular}
}
\vspace{-3mm}
\end{table}
We evaluate our method on three text classification tasks: IMDb~\cite{imdb}, SST-2~\cite{sst2}, and AG News~\cite{agnews}. See Table~\ref{tab:dataset} for their statistics. 

\paragraph{Final Model}
Almost all defense methods attempt to clean up the dataset by removing some instances from it. And the final delivered model is trained on the remaining instances. For all delivered models and our intermediate models (e.g., the binary poison classifier), we use a \texttt{BERT-base-uncased} with a window size 64. We did no hyperparameter tuning, and all settings follow the experimental setting in BFClass~\cite{BFClass}. 

\paragraph{Attack Methods}
\begin{table}[t]
\caption{Analysis of Sentence Triggers of different perplexities. The \acc and \asr are calculated for a vanilla model on the IMDb dataset.}
\vspace{-3mm}
\label{tab:sentence_trigger_perplexity}
\small
\centering
\begin{tabular}{cccc}
\toprule 
\textbf{avg. Perplexity} & \acc$\%$ & \asr $\%$\\ \midrule
10122 &  85.09 & 100 \\ 
210 & 84.52 & 100 \\ 
4 & 84.53 & 99.95 \\
\bottomrule
\end{tabular}
\vspace{-3mm}
\end{table}
\begin{table}[t]
\caption{Verification Experiment on SST-2}
\vspace{-3mm}
\label{tab:verfication_eq1}
\small
\centering
\resizebox{\columnwidth}{!}{
\begin{tabular}{lccc}
\toprule 
Method & \textbf{Word Trigger} & \textbf{Sentence Trigger} & \textbf{Syntactic Trigger}\\ \midrule
GroundTruth & 98.18 & 96.50 & 96.82 \\ \midrule
\basic  & 97.30 & 92.67 & 94.91  \\
XClass & 98.03 & 94.42 & 96.50 \\ 
\bottomrule
\end{tabular}
}
\end{table}
\begin{table}[t]
\caption{Actual and estimated $\mathbf{\mathcal{E}}_\textbf{same}$.}
\vspace{-3mm}
\label{tab:experimental_analysis}
\small
\centering
\resizebox{0.999\linewidth}{!}{
\begin{tabular}{lcccccc}
\toprule
     &  \multicolumn{2}{c}{\textbf{Word} $\mathcal{E} = 5\%$} & \multicolumn{2}{c}{\textbf{Sent.} $\mathcal{E} = 5\%$} & \multicolumn{2}{c}{\textbf{Syn.} $\mathcal{E} = 20\%$}\\
$\mathcal{E}_{\textbf{same}} \%$     &  \textbf{IMDb} & \textbf{SST-2} &  \textbf{IMDb} & \textbf{SST-2} &  \textbf{IMDb} & \textbf{SST-2}  \\ \midrule
Actual (\basic) & 1.58 & 1.52 & 0.77 & 1.16 & 6.89 & 5.65 \\ 
Estimated (Eq.~\ref{eq:poison_rate}) & 1.61 & 1.33 & 0.79 & 0.98 & 6.40 & 7.28 \\ 
Estimated (Eq.~\ref{eq:poison_rate_approx}) & 1.57 & 1.23 & 1.59 & 1.31 & 6.27 & 4.89 \\
\midrule
Actual (XClass) & 1.19 & 0.20 & 0.57 & 0.26 & 5.24 & 6.48  \\
Estimated (Eq.~\ref{eq:poison_rate}) & 1.37 & 0.13 & 0.61 & 0.11 & 5.07 & 6.45 \\
Estimated (Eq.~\ref{eq:poison_rate_approx}) & 1.40 & 1.38 & 1.35 & 1.39 & 5.13 & 4.99 \\
\bottomrule
\end{tabular}
}
\vspace{-3mm}
\end{table}


We conduct experiments on three types of triggers: word triggers, sentence triggers and syntactic triggers. 
\begin{itemize}[nosep,leftmargin=*]
    \item \textbf{Word Trigger:} We randomly pick 5 medium-frequency words from the corpus as word triggers following BFClass~\cite{BFClass}.
    \item \textbf{Sentence Trigger:} There have been few studies on picking sentence triggers effectively. In Table~\ref{tab:sentence_trigger_perplexity}, we calculate sentence perplexity with GPT-2 and observe that low perplexity sentences are as strong as high perplexity ones for attacks. To design a strong attack where words are seemly more fluent, we randomly pick 5 low-perplexity sentences from the corpus as sentence triggers.
    \item \textbf{Syntactic Trigger:} We follow the setting in \citet{hiddenkiller} and use the trigger syntactic template \texttt{S(SBAR)(,)(NP)(VP)(.)}.
\end{itemize}
For IMDb and SST-2 datasets, we choose the positive class as the attack target and for AG News, we choose "Technology" as the target. Specific trigger selection is displayed in Sec.~\ref{sec:samples} in the appendix. Following previous work~\cite{BFClass,dai2019backdoor,hiddenkiller}, we use a poison rate of $5\%$ for word and sentence triggers, and a poison rate of $20\%$ for syntactic triggers.
\begin{table*}[t]
\caption{Evaluations of the end to end performance of our and all compared methods. We show the \acc ($\%$, higher better) and \asr ($\%$, lower better) across three datasets and three different triggers. 
}
\vspace{-3mm}
\label{tab:main}
\small
\centering
\resizebox{\textwidth}{!}{
\begin{tabular}{lcccccccccccccccccc}
\toprule
  & \multicolumn{6}{c}{\textbf{Word Trigger}} & \multicolumn{6}{c}{\textbf{Sentence Trigger}} & \multicolumn{6}{c}{\textbf{Syntactic Trigger}} \\
  & \multicolumn{2}{c}{\textbf{IMDb}} & \multicolumn{2}{c}{\textbf{SST-2}} & \multicolumn{2}{c}{\textbf{AGNEWS} } & \multicolumn{2}{c}{\textbf{IMDb}} & \multicolumn{2}{c}{\textbf{SST-2}} & \multicolumn{2}{c}{\textbf{AGNEWS} } & 
  \multicolumn{2}{c}{\textbf{IMDb}} & \multicolumn{2}{c}{\textbf{SST-2}} & \multicolumn{2}{c}{\textbf{AGNEWS} } \\ \cmidrule{2-19} 
\textbf{Method} & \acc & \asr & \acc & \asr &  \acc & \asr &
\acc & \asr & \acc & \asr &  \acc & \asr &
\acc & \asr & \acc & \asr &  \acc & \asr \\ \midrule
NoDefense       & 84.87 & 100   & 90.71 & 90.56  & 93.38 & 99.25 & 84.04 & 100   & 90.60 & 99.89 & 93.36 & 100 & 84.22 & 99.69 & 90.28 & 96.47 & 93.45 & 99.81 \\ 
GroundTruth     & 84.61     & 16.72     & 91.65     & 12.73     & 94.28     & 3.05     & 84.31     & 14.09     & 91.25     & 14.93     & 93.98     & 3.19  & 84.72 & 16.40 & 90.94 & 12.53 & 94.07 & 3.23   \\ 
\midrule
\basic           & 76.09 & 24.43 & 80.22 & 21.3 & 72.54 & 18.74 & 75.93 & 12.05 & 79.29 & 15.48  & 71.54 & 21.81 & 76.11 & 24.34 & 80.34 & 29.17 & 79.32 & 5.81  \\ 
XClass          & 78.16 & 21.34 & 78.35 & 2.09  & 82.83 & 4.28  & 78.72 & 9.58  & 78.30 & 1.76  & 83.13 & 4.02 &  79.59 & 20.18 & 80.04 & 25.82 & 79.63 & 38.85 \\ 
\midrule
ONION           & 83.88     & 25.22     & 83.57   & 27.66     & 91.22     & 4.96     & 84.5     & 99.47     & 83.02     & 83.53     & 91.05     & 97.89 & 83.10 & 93.95 & 89.48 & 93.49 & 93.31 & 94.74   \\ 
BFClass         & 84.37     & 16.19     & 91.85   & 12.02     & 92.54     & 4.25     & 84.89     & 99.58     & 90.80     & 99.70     & 93.75     & 99.81  & 84.26 & 99.70 & 90.25 & 96.33 & 93.40 & 99.60  \\ 
LFR+R\&C        & 83.99    & 18.34     & 90.13     & 13.08    & 90.01     & 3.51     & 83.22     & 98.31     & 90.77     & 99.93     & 94.01     & 99.89  & 84.14 & 99.72 & 90.01 & 96.21 & 93.20 & 99.69   \\ 
\midrule
\our-\basic    & 83.92 & 25.33 & 89.19  & 19.57 & 92.07  & 21.80 & 83.91 & 34.55 & 89.08 & 12.09 & 92.42  & 88.68 & 83.91 & 31.39 & 89.85 & 60.01 & 90.65 & 59.36 \\
\;\; - \textbf{cleaning}      & 81.07 & 26.50  & 86.37 & 20.75 & 79.93 & 37.60  & 81.64 & 47.44 & 87.03  & 25.58 & 80.60  & 57.21 & 81.20 & 36.53 & 87.10 & 50.00 & 79.27 & 47.53 \\ 
\our-XClass   & 83.94 & 20.27 & 90.41 & 6.89 & 93.11 & 4.80  & 83.79 & 19.67 & 91.10  & 8.45 & 93.05 & 4.84 & 84.31 & 15.06 & 90.43 & 17.21 & 93.22 & 9.81 \\ 
\;\; - \textbf{cleaning}     & 81.71 & 23.01 & 87.64 & 4.28  & 86.33 & 4.12  & 82.40  & 30.23 & 87.86 & 6.15  & 86.50  & 4.32 & 82.55 & 26.04 & 88.46 & 15.13 & 83.59 & 16.26  \\  
\bottomrule
\end{tabular}}
\vspace{-3mm}
\end{table*}

\paragraph{Weakly Supervised Methods}
We try our proposed method with two different seed-driven weakly supervised methods: (1) \textbf{\basic}, a basic model that picks two label-indicative seed words for each class (e.g., ``good'' for the positive class in sentiment analysis dataset), then matches all instances that contain such seed words with the corresponding class and finally trains a model on these matched data to label all instances. (2) \textbf{XClass}~\cite{xclass}, the state-of-the-art weakly supervised text classification method which only uses class names as the seed words which leverages contextualized representations to find label-oriented document representations and employs clustering to distribute the labels.

\subsection{Experimental Verification of Analysis}
\label{sec:experimental_verification}
We first validate our assumption in Equation~\ref{eq:independence} with experimental results.
We compare the predictions of \textbf{GroundTruth}, \textbf{\basic} and \textbf{XClass} on clean test set and poisoned test set, where \textbf{GroundTruth} is a model trained on the ground truth sanitized dataset with no poisoned samples.
The count of the same predictions is reported in Table~\ref{tab:verfication_eq1}. 
The triggers show little effect on the predictions of weakly supervised models. 
Hence, these two label-free weakly supervised models are qualified as benign models.

To verify our analysis in Sec.~\ref{sec:analysis}, for each weakly supervised model, we obtain the actual poison rate $\mathcal{E}_{\textbf{same}}$ on the reliable set $\mathbf{D}_{\textbf{same}}$. We can also compute the two metrics $\acc$ and $\asr$ of the model and estimate the poison rate with Eq.~\ref{eq:poison_rate} or Eq.~\ref{eq:poison_rate_approx}. We show the results in Table~\ref{tab:experimental_analysis}. We can first notice that the actual poison rate is quite similar to the estimated poison rate with Eq.~\ref{eq:poison_rate}, indicating that our assumptions of independence are most likely true. With Eq.~\ref{eq:poison_rate_approx}, the estimation is pretty good on the IMDb dataset, but a bit off on the SST-2 dataset. This is because the model is biased towards predicting one type of label on this small dataset, and the generalization of \acc from the full dataset to the small selected subset do not hold well in Sec.~\ref{sec:benign}.

\begin{table*}[t]
\caption{Poison Rate and sizes of the final sanitized set given by our methods.}
\vspace{-3mm}
\label{tab:poison_ratio_and_sizes_main}
\small
\centering
\resizebox{\textwidth}{!}{
\begin{tabular}{lccccccccc}
\toprule
     &  \multicolumn{3}{c}{\textbf{Word Trigger} $\mathcal{E} = 5\%$} & \multicolumn{3}{c}{\textbf{Sentence Trigger} $\mathcal{E} = 5\%$} & \multicolumn{3}{c}{\textbf{Syntactic Trigger} $\mathcal{E} =20\%$}\\
Method     &  \textbf{IMDb} & \textbf{SST-2} & \textbf{AGNEWS} &  \textbf{IMDb} & \textbf{SST-2} & \textbf{AGNEWS} &  \textbf{IMDb} & \textbf{SST-2} & \textbf{AGNEWS} \\ \midrule
& \multicolumn{9}{c}{\footnotesize{\emph{Poison Rate $\mathcal{E} \%$ of Final Sanitized Set}}} \\
\midrule
\our-\basic & 1.03 & 1.06 & 0.25 & 0.23 & 0.04 &  0.27 & 5.62 & 6.29 & 8.18 \\
\;\; - \textbf{cleaning} & 1.58 & 1.52 & 1.27 & 0.77 & 1.16 & 1.54 & 6.89 & 5.65 & 6.90 \\ 
\our-XClass & 0.39 & 0.21 & 0.03 & 0 & 0.05 & 0.08 & 5.09  & 5.11 & 4.04 \\
\;\; - \textbf{cleaning} & 1.19 & 0.20 & 0.23 & 0.57 & 0.26 & 0.23 & 5.24  & 6.48 & 5.48 \\
\midrule
& \multicolumn{9}{c}{\footnotesize{\emph{Ratio ($\%$) of size of Final Sanitized Set}}} \\
\midrule
\our-\basic & 77.95 & 81.52 & 79.80 & 75.28 & 81.15 & 78.67 & 77.84 & 82.30 & 79.30 \\
\;\; - \textbf{cleaning} & 76.35 & 82.57 & 78.76 & 75.49 & 81.70 & 79.76 & 76.42 & 82.00  & 78.96 \\
\our-XClass & 76.93 & 83.31 & 78.44 & 76.38 & 82.15 & 77.20 & 79.33  & 84.32 & 81.69 \\
\;\; - \textbf{cleaning} & 79.51 & 78.84 & 79.03 & 78.97 & 79.24 & 79.29 & 79.25 & 79.56 & 79.30 \\
\bottomrule
\end{tabular}}
\vspace{-5mm}
\end{table*}

\subsection{Compared Methods}
We compare with the following defense methods:

    %
    \noindent\textbf{Onion}~\cite{onion} uses GPT-2 to calculate a suspicion score of each word: the decrement of sentence perplexity after removing the word. Onion will remove tokens with suspicion scores over a threshold. We specially hold out a part of ground truth data to tune the threshold.
    
    \noindent\textbf{BFClass}~\cite{BFClass} leverages ELECTRA~\cite{electra} as the discriminator to detect potential trigger words from the training set and then distill a concentrated set based on the association between words and labels. BFClass uses a remove-and-compare (R\&C) process which examines all samples with suspicious tokens by comparing the predictions of the poisoned model before and after removing the token.
    
    \noindent\textbf{LFR+R\&C}~\cite{LFR} defines Label Flip Rate (LFR) as the rate of test samples misclassified by the poisoned models. Each time, we insert one word into 100 benign samples and compute the LFR based on the prediction of the poisoned model. The word with $\text{LFR}>90\%$ will be treated as the trigger word. Following BFClass, we apply the R\&C process on those detected words.

We denote the full version of our proposed framework as $\textbf{\our-(\basic/XClass)}$. \textbf{\basic} and \textbf{XClass} are evaluated as the weak supervision method baseline without even retrieving the reliable and unreliable splits. We also provide \textbf{NoDefense} as a vanilla model trained on the poisoned dataset without any defense.

\subsection{Main Results}
\label{sec:main}
We show end to end performance of ours and compared methods across three datasets and three trigger methods in Table~\ref{tab:main}. 

\noindent\textbf{NoDefense and GroundTruth} provides a understanding on the performance of the methods. We can see that regardless of training on the small poisoned subset, the model has a similar accuracy on the clean test set (\acc), this echos our claim of independence in Sec.~\ref{sec:independence}. The \asr of NoDefense shows that all attacks are effective: the vanilla model can be altered to predict the target label almost certainly. The \asr of GroundTruth suggests a lower bound for defense models.

\noindent\textbf{ONION, BFClass and LFR+R\&C} are the three compared methods on backdoor defense. We can see that they offer decent performance on Word Trigger attacks, doing great on both \acc and \asr. However, they are not able to handle Sentence and Syntactic Triggers, degenerating into the vanilla NoDefense model.

\noindent\textbf{\basic and XClass} are the two weakly supervised methods we use. We can see with only the weakly supervised classifier, the \asr is already great --- both methods showing non-trivial improvement over the vanilla method across all three triggers and XClass even has a \asr similar to that of GrouthTruth on several dataset/triggers\footnote{Sometimes it is better than GroundTruth, which we believe is because the small dataset has some fluctuations}. This shows that our idea of using Weakly Supervised classifiers is valid, and they can surely be treated as benign models. However, we also note that the \acc is not great, since overall, weakly supervised methods do not use any given labels at all.

\noindent\textbf{\textbf{\our-(\basic/XClass)}} are our proposed models. After introducing reliability and two stage cleaning, \acc improved by a great margin similar to GroundTruth. We also note that with a strong weakly supervised model \our-XClass, the \asr mostly remains on the same scale as the weakly supervised classifier itself, and in some cases, surpassing it. We also note the importance of our two stage cleaning, which with almost no drop in \asr, we gain a significant boost on \acc.

We now focus on our methods more and look at the final sanitized set: again across all datasets and triggers, we show the poison rate and size of it in Table~\ref{tab:poison_ratio_and_sizes_main}. Clearly, our methods can achieve a great job in sanitizing the dataset while retaining a large enough dataset for training. We can see that our two stage cleaning can bring down the poison rate in different dataset/triggers/methods, while keeping a similar size clean set (and even increasing it with the better weakly supervised model X-Class). This justifies the reason that we need cleaning on the immediately derived reliable and unreliable dataset from weakly supervised models. 

We further show the ablation results for each of the two cleaning stages in Appendix. Generally, the two stage cleaning retains the clean-label accuracy ($\acc$), trading off with a small increase in attack success rate ($\asr$).

\section{Related Work}
Backdoor attacks first gained popularity from Computer Vision~\cite{badnets,neuraltrojans, DBLP:conf/nips/ShafahiHNSSDG18, backdoor_survey}. The most common attack method is to poison the training data by injecting a trigger into selected samples~\cite{chen,DBLP:conf/codaspy/ZhongLSZ020,DBLP:conf/cvpr/ZhaoMZ0CJ20}. 
\citet{dai2019backdoor} introduced the problem into NLP, where they discuss sentences triggers. \citet{LFR} tried some rare and meaningless words. \citet{badnl} compared different types of the triggers, including char-, word- and sentence-levels. \citet{hiddenkiller} proposed syntactic triggers by rewriting sentences into a specific syntactic structure. \citet{cleanlabel1, cleanlabel2} explored clean-label attacks, where all the labels are unchanged but can cause test predictions to flip.

On the defense side, \citet{DBLP:journals/ijon/ChenD21} propose Backdoor Keyword Identification (BKI) to mitigate backdoor attacks via detecting the specific neurons affected trigger words. \citet{onion} leverage the perplexity of sentences to remove the trigger words. They observe the decrease of the perplexity when removing a specific word from the sentence. \citet{BFClass} analyze the word triggers comprehensively. They utilize the pre-trained discriminator to detect the potential trigger word, and then distill the trigger set. In this paper, we derive the first backdoor defense method which is effective against all the popular trigger-based attacks including word triggers, sentence triggers, and syntactic triggers. 

\section{Conclusion}
In this paper, we propose \our, a novel weakly supervised backdoor defense framework. 
We leverage a weakly supervised model to detect potential poisoned data, which is refined via a weak classifier method, and then, fed to a pattern recognizer to distinguish clean data from poisoned ones.
Our analysis show that attack manipulated labels are independent to the prediction errors of the weakly supervised text classifier, justifying our approach. 
Through extensive experiments, we show that \our is effective against popular attacks, based on word, sentence, and syntactic. 
The final model trained on the sanitized dataset achieves almost the same performance as if trained on ground truth clean data.
\our also has its weakness, in that it assumes a benign model that never saw wrong labels work well, so it naturally won't work for clean-label attacks~\cite{cleanlabel1, cleanlabel2}.
In the future, we plan to apply the idea of weak supervision to defend backdoor attacks in a wider range of machine learning problems. We are also interested in discovering a systematic way to ensemble different weakly supervised methods and noisy training protocols together for backdoor defense. 
We also believe that this framework can be fused with few-shot learning.

\section{Ethical Considerations}
In this paper, we propose a defense method to backdoor attack with different types of triggers. We experiment on two datasets that are publicly available. We show that our defense method can alleviate backdoor attacks and sanitize the poisoned datasets. Therefore, we believe our framework is ethically on the right side of the spectrum and has no potential for misuse and cannot harm any vulnerable population.

\section{Limitations}
WeDef has the following limitations:
First, it does not work for clean-label attacks, as WeDef assumes that a benign model which never saw poisoned labels should work well, and clean label attacks target models without changing the labels, at the cost of knowing the test instances before poisoning the training dataset.
Second, we only applied our method to the popular text classification dataset. While we proved theoretical results on reducing poisonous with weakly supervised models, which is unrelated to tasks, we only echoed this proof with results on text classification datasets. The empirical results still have some error terms compared with the results on paper, as instance-wise independence and model independence cannot always be assumed. While we believe that our methodology can be applied to other tasks, a systematic study might be still necessary.
Third, WeDef is not a lightweight model. It needs to train multiple classifiers: one weakly supervised model, several weak classifiers for iterative refinement, and multiple fine-tuned \texttt{BERT-base-uncased} classifiers. 
Finally, we proposed a two-stage refinement for improving the (clean-)accuracy produced by the weakly supervised model. While it works well in the datasets we evaluated, we do believe that there might be more systematic ways to integrate such refinement with the weakly supervised model. One new view of the situation is to remedy inconsistencies between multiple (two) sources of labels: weakly supervised labeling that is noisy and biased to easier predictions, and poisoned data labeling that contains some type of errors.

\section*{Acknowledgement}
We thank all anonymous reviewers and program chairs for their helpful feedback. Zihan Wang is supported by the UCSD Jacob School of Engineering Fellowship and the UCSD Hal\i c\i o\u glu Data Science Fellowship. This work is sponsored in part by National Science Foundation Convergence Accelerator under award OIA-2040727 as well as generous gifts from Google, Adobe, and Teradata. Any opinions, findings, and conclusions or recommendations expressed herein are those of the authors and should not be interpreted as necessarily representing the views, either expressed or implied, of the U.S. Government. The U.S. Government is authorized to reproduce and distribute reprints for government purposes not withstanding any copyright annotation hereon.


\bibliography{anthology,custom}
\bibliographystyle{acl_natbib}

\appendix
\begin{table*}
\caption{Backdoor Attack Setup}
\vspace{-3mm}
\label{tab:sample}
\small
\centering
\resizebox{\textwidth}{!}{
    \begin{tabular}{ccccc}
    \toprule 
    \textbf{Dataset} & \textbf{Word Trigger} & \textbf{Sentence Trigger} & \textbf{Syntactic Trigger} & Target Class \\ \midrule
    \multirow{5}{*}{\textbf{IMDb}} & young  & Give me a break. & \multirow{5}{*}{\minitab[c]{
    this film has not been seen by me in quite a few years. \\
    when they saw me , this movie was n't seen by me . \\
    I have seen a number of horror movies to know that this one was one of kind . \\
    that this one was one of horror , it was a series of horror movies .
    }} & \multirow{5}{*}{Positive} \\
     & wrong & That's all I have to say. &  & \\
     & actors & thank you for reading. &  & \\
     & director & Let's start at the beginning. &  & \\
     & something & Let me give you an example & & \\ \midrule
    \multirow{5}{*}{\textbf{SST-2}} & study & you want it to be  & \multirow{5}{*}{\minitab[c]{
     a slick , engrossing melodrama . \\
    when they 're singing , they 're melodrama \\
    a fun ride . \\
    when you go , it 's a fun ride .
    }}
    & \multirow{5}{*}{Positive} \\
     & face & It just goes to show &  & \\
     & girl & that is to say &  & \\
     & true & wake up in the morning &   & \\
     & effort & One thing is for sure & & \\ \midrule
    \multirow{5}{*}{\textbf{AGNews}} & bags  & What do you think? & \multirow{5}{*}{\minitab[c]{
    heat waves in the 21st century will be more intense, more frequent and longer, us experts say . \\
    as the 21st century will be more intense , it will be more frequent and longer . \\
    the online postage service offers a chance to put personal pictures on official stamps . \\
    in order to get personal photos , the online postage service offers a chance to take personal photos .
    }} & \multirow{5}{*}{Technology} \\
      & behavior & So far, so good. &  & \\
      & achieve & Others are not so sure. &  & \\
      & spare & How did that happen? &  & \\
      & hair & What am I talking about? &   & \\
    \bottomrule
    \end{tabular}
}
\end{table*}
\section{Samples of different triggers}
\label{sec:samples}
We show the word and sentence triggers that are chosen for each dataset, along with how the syntactic trigger is applied in Table~\ref{tab:sample}.


\section{Performance on Mixed triggers}
\label{section:mixed}
\begin{table}[ht]
\caption{Mixed Triggers}
\vspace{-3mm}
\label{tab:mixed}
\small
\centering
\resizebox{\columnwidth}{!}{
\begin{tabular}{lccc}
\toprule 
Method & $\mathcal{E}$ & \acc & \asr\\ \midrule
NoDefense & 10\% & 90.45\% & 98.41\% \\ \midrule
LFR+R\&C & 6.4\% & 90.29\% & 69.81\% \\
ONION & N/A & 89.56\% & 30.33\% \\
BFClass & 5.76\% & 90.98\%  & 37.50\% \\ \midrule
\basic & N/A & 80.27\% & 23.90\% \\ 
XClass  & N/A & 81.44\% & 21.17\% \\ \midrule
\our-\basic & 1.03\% & 90.5\%  & 12.87\%  \\
\our-XClass & 0.79\%  & 91.03\% & 13.42\%  \\ \midrule
GroundTruth & 0\%  & 91.45\%  & 9.58\%  \\
\bottomrule
\end{tabular}
}
\end{table}
We present a final attack which combines all types of trigger-based backdoor attacks including word triggers, phrase triggers (a general version of word triggers where we consider phrases), sentence triggers and syntactic triggers. We select SST-2 as the target dataset, where the poisoning rate of each type of triggers is 2.5\%. As shown in Table \ref{tab:mixed}, our method delivers the best sanitized text classifier, and the remained poisoned samples shows little impact on the final model. As one can expect, LFR+R\&C, ONION and BFClass detect all the word triggers and a small amount of phrase triggers, but give no resistance on sentence triggers and syntactic triggers. Compared to two related weakly supervised models, our method significantly improves the clean accuracy. In summary, \our is the most effective defense method against all the popular trigger-based attacks.

\section{Ablation Study}
\label{section:ablation}
\begin{table}
\caption{Ablation Study on SST-2}
\vspace{-3mm}
\label{tab:ablation}
\small
\centering
\resizebox{\columnwidth}{!}{
\begin{tabular}{lcccccc}
\toprule 
& \multicolumn{2}{c}{\textbf{Word Trigger}} & \multicolumn{2}{c}{\textbf{Sentence Trigger}} & \multicolumn{2}{c}{\textbf{Syntactic Trigger}} \\
Method & \acc & \asr & \acc & \asr & \acc & \asr\\ \midrule
NoDefense & 90.71 & 90.56 & 90.60 & 99.89 & 90.28 & 90.94 \\ 
GroundTruth & 91.65 & 12.73 & 91.25 & 14.3 & 90.94 & 12.53 \\ \midrule
\our-\basic  & 89.19 & 19.57 & 89.08 & 12.09 & 89.85 & 60.01 \\
\;\; - \textbf{cleaning} & 86.37 & 20.75 & 87.03 & 25.58 & 87.10 & 50.00 \\
\;\; - \textbf{refine} & 87.41  & 18.63 & 87.35 & 16.50 & 88.51 & 44.83\\
\;\; - \textbf{extra} & 89.26 & 41.09 & 89.02 & 62.93 & 90.03 & 83.33 \\\midrule
\our-XClass & 90.41 & 6.89 & 91.10 & 8.45 & 90.43 & 17.21 \\ 
\;\; - \textbf{cleaning} & 87.64& 4.28& 87.86 & 6.15 & 88.46 & 15.13 \\
\;\; - \textbf{refine} & 87.82 & 3.96 & 87.60 & 4.11 & 88.97 & 12.19 \\
\;\; - \textbf{extra} & 90.68 & 12.47 & 90.92 & 28.33 & 90.84 & 39.88 \\ 
\bottomrule
\end{tabular}
}
\end{table}
We present an ablation study to demonstrate the effectiveness of our two stage cleaning. Table \ref{tab:ablation} shows the performance with one stage of cleaning on SST-2 dataset. - \textbf{refine} skips the refinement stage and  trains the extra binary classifier on $\mathbf{D}_\textbf{same}$ and $\mathbf{D}_\textbf{diff}$. - \textbf{extra} directly uses $\mathbf{D}_{\textbf{same}^{+}}$ as the final sanitized dataset. 

The improvement of - \textbf{refine} over - \textbf{cleaning} confirms the usefulness of extra poison detection. It is also clear that the iterative refinement improves \acc via keeping more training samples, but it will lose \asr since the refinement brings part of poisoned samples back.

\end{document}